\pdfoutput=1
\documentclass[conference]{IEEEtran}
\IEEEoverridecommandlockouts
\usepackage{amsmath,amssymb,amsfonts}
\usepackage{algorithmic}
\usepackage{graphicx}
\usepackage{textcomp}
\usepackage{xcolor}
\usepackage{multirow}
\usepackage{blindtext}
\usepackage{tabularx}
\usepackage{mathtools, commath}
\usepackage[sorting=none]{biblatex}
\usepackage{stfloats}
\usepackage{caption}
\usepackage{float}
\usepackage[section]{placeins}
\usepackage{mathtools,amssymb}
\makeatletter
\AtBeginDocument{%
  \expandafter\renewcommand\expandafter\subsection\expandafter{%
    \expandafter\@fb@secFB\subsection
  }%
}
\makeatother
\def\BibTeX{{\rm B\kern-.05em{\sc i\kern-.025em b}\kern-.08em
    T\kern-.1667em\lower.7ex\hbox{E}\kern-.125emX}}
\begin{document}

\author{
\IEEEauthorblockN{
Sideshwar J B\IEEEauthorrefmark{1},
Sachin Krishan T\IEEEauthorrefmark{1},
Vishal Nagarajan\IEEEauthorrefmark{1},
Shanthakumar S\IEEEauthorrefmark{2},
Vineeth Vijayaraghavan\IEEEauthorrefmark{2}
}
\IEEEauthorblockA{
\IEEEauthorrefmark{1}SSN College of Engineering, Chennai, India\\
\IEEEauthorrefmark{2}Solarillion Foundation, Chennai, India\\
\texttt{\{sideshwar18151, sachinkrishnan18128, vishal18198\}@cse.ssn.edu.in}\\
\texttt{shantha2106@gmail.com}\\
\texttt{vineethv@ieee.org}
}
}

\title{End-to-End Optimized Arrhythmia Detection Pipeline using Machine Learning for Ultra-Edge Devices\\}

\maketitle

\begin{abstract}
Atrial fibrillation (AF) is the most prevalent cardiac arrhythmia worldwide, with 2\% of the population affected. It is associated with an increased risk of strokes, heart failure and other heart-related complications. Monitoring at-risk individuals and detecting asymptomatic AF could result in considerable public health benefits, as individuals with asymptomatic AF could take preventive measures with lifestyle changes. With increasing affordability to wearables, personalized health care is becoming more accessible. These personalized healthcare solutions require accurate classification of bio-signals while being computationally inexpensive. By making inferences on-device, we avoid issues inherent to cloud-based systems such as latency and network connection dependency. We propose an efficient pipeline for real-time Atrial Fibrillation Detection with high accuracy that can be deployed in ultra-edge devices. The feature engineering employed in this research catered to optimizing the resource-efficient classifier used in the proposed pipeline, which was able to outperform the best performing standard ML model by $10^5\times$ in terms of memory footprint with a mere trade-off of 2\% classification accuracy. We also obtain higher accuracy of approximately 6\% while consuming 403$\times$ lesser memory and being 5.2$\times$ faster compared to the previous state-of-the-art (SoA) embedded implementation.
\end{abstract}


\section{Introduction}
Fibrillation, the most serious form of arrhythmia, is fast, uncoordinated beats, which are contractions of individual heart-muscle fibers. Atrial fibrillation (also called AFib or AF) is the most common type of supraventricular arrhythmia which is defined as a tachyarrhythmia characterized by predominantly uncoordinated atrial activation with consequent deterioration of atrial mechanical function \cite{1} and can be chronic. AF is the most common and sustained cardiac arrhythmia, occurring in 1-2\% of the world population \cite{2}, \cite{3} and prevalence of AF in the general population is expected to increase over the coming years due to an aging population. AF is associated with a 5-fold increased risk of ischemic stroke, 3-fold increased risk of heart failure, and 2-fold increased risk of heart disease-related death \cite{4, 5, 6}. According to a survey by Sumeet et al. \cite{7} more than 12 million Europeans and North Americans are estimated to suffer from AF and its prevalence will likely triple in the next 30-50 years \cite{5}. More importantly, the incidence of AF increases with age, from less than 0.5\% at 40-50 years of age to 5-15\% for 80 years old \cite{6}.

A person suffering from AF may exhibit symptoms like palpitations – sensations of racing, irregular heartbeat or flip-flopping in the chest, reduced ability to exercise, fatigue, dizziness, shortness of breath, and chest pain. However, for at least one-third of the patients, it is intermittent and silent. As a result, stroke, heart failure, and death can be the first presenting symptom of someone who has asymptomatic AF. This emphasizes the need for AF surveillance methods, which would allow early diagnosis and treatment.

While standard Machine Learning and Deep Learning models perform exceptionally well in the field of detection, the computation cost associated with it is notoriously high to be deployed on ubiquitous wearables. One solution to tackle this problem is by using cloud servers for computations that have virtually no limit on computational power. Albeit enabling usage of SoA models, there are two problems accompanying this approach including high latency in real-time applications and privacy concerns over sensitive healthcare data. This calls for on-device (offline) AF detection in low-power microcontrollers/microprocessors present in ultra-edge devices and wearables.

This research proposes a resource-efficient inference pipeline for accurate AF detection. The pipeline is optimized for the least inference time and is deployable on low-power embedded devices. This was made possible by the model’s meager memory requirements. The robustness of the proposed architecture was validated for AF detection by training it on one dataset and testing on another dataset to emulate real-world clinical scenarios. To appraise the performance of the architecture for general arrhythmia detection, it was tested to detect Ventricular Fibrillation (VF).

This paper is structured as follows, Section \ref{Existing Work} talks about the existing work, their respective approaches and their results, Section \ref{Datasets} briefly talks about the datasets used in this research. Section \ref{Dataset preprocessing} elaborates on the data preprocessing. Section \ref{Pipeline Arch} explains the pipeline architecture. Section \ref{Feature Selection} elucidates about feature importance and training employed in this research. Section \ref{Results and Analysis} provides a detailed analysis of the results.

\section{Related Work}
\label{Existing Work}
In order to perform AF detection, several machine learning and deep learning approaches exist. Y. Jin et al. \cite{8} developed a novel Domain Adaptive Residual Network (DARN) and was able to achieve sensitivity - 98.97\%, specificity - 98.75\% and accuracy - 98.84\% for the MIT-BIH Atrial Fibrillation DataBase (AFDB). Jibin Wang et al. \cite{9} constructed 11-layer network architecture to automatically classify AF and AFL (Atrial Flutter) signals and achieved accuracy, specificity and sensitivity of 98.8\%, 98.6\% and 98.9\% respectively on the AFDB dataset. Sajad Mousavi et al \cite{10} developed an attention network model. Their proposed method achieved the SoA scores on AFDB with an accuracy of 99.40\%, sensitivity of 99.53\% and specificity of 99.26\%. Although there exists considerable research in the field of AF detection with exceptional performances, these solutions are computationally expensive and cannot be used to perform on-device predictions on resource-constrained devices.

Cloud-based approaches were proposed by Nurul Huda et al. \cite{11} and Samuel A Setiawan et al. \cite{12} to tackle the lack of computational power in wearable devices. However, high operational latencies due to data transmission in cloud-based approaches adversely affect real-time monitoring systems. To overcome this limitation, Antonino Faraone et al. \cite{13} proposed a solution for AF detection on resource-constrained devices. They implemented an optimized Convolutional-Recurrent Neural Network for ECG-based detection of arrhythmias. They deployed their proposed model on an embedded device and were able to achieve an accuracy of 86.1\%, memory footprint of 210 KB and inference time of 94.8ms.

Our research proposes and implements a more efficient and clinically applicable inference pipeline for real-time AF detection. This pipeline uses a lightweight machine learning (ML) classifier to detect AF which occupies memory in the order of 0.5KB and processing time of a few milliseconds.

\section{Datasets}
\label{Datasets}

MIT-BIH Atrial Fibrillation DataBase (AFDB) \cite{14} and Computing in Cardiology Challenge 2017 Database (2017 Challenge Dataset) \cite{15} were considered for AF detection. Ventricular Fibrillation DataBase (VFDB) \cite{16} was chosen for VF detection. All the datasets discussed below are available in PhysioNet \cite{17}. 

\subsection{AFDB}\label{AA}
This dataset \cite{14} includes 25 long-term ECG recordings of human subjects with atrial fibrillation. Out of the 25 records, two of them were devoid of usable data and hence they were not used in this research. The remaining 23 records contain two ECG signals each. The individual recordings are each 10 hours in duration and contain two ECG signals each sampled at 250 Hz with 12-bit resolution over a range of ±10 millivolts. The original analog recordings were made at Boston's Beth Israel Hospital using ambulatory ECG recorders with a typical recording bandwidth of approximately 0.1 Hz to 40 Hz.

\subsection{2017 Challenge Dataset (CHDB)}
CHDB \cite{15} contains 2 sets of ECG recordings. The training set consists of 8528 recordings and the testing set consists of 3658 recordings lasting from 9 seconds to 31 seconds. The ECG data were recorded using the AliveCor device. The data were digitized in real-time at 44.1 kHz and 24-bit resolution using software demodulation. Finally, the data was stored at 300 Hz.

\subsection{VFDB}
This dataset \cite{16} consists of 30 minutes recordings of 22 subjects who experienced episodes of sustained ventricular tachycardia, ventricular flutter and ventricular fibrillation digitized at 250 Hz. Unlike the AF datasets, VFDB does not provide QRS annotations. 

\section{Dataset preprocessing}
\label{Dataset preprocessing}

\begin{figure}[h]
\centerline{\includegraphics[trim=0 0 0 1cm,width=0.9\columnwidth,clip]{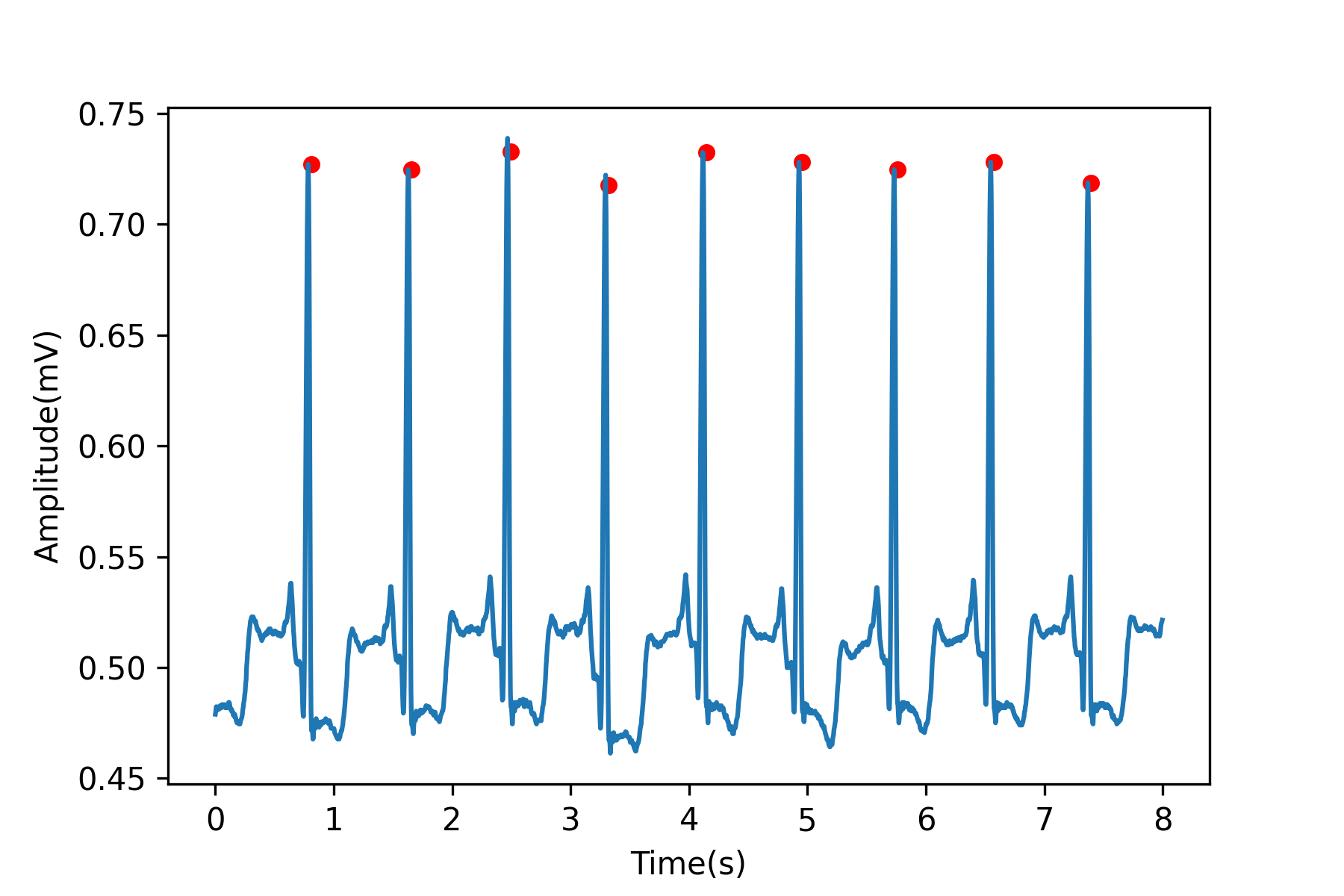}}
\captionsetup{belowskip=-10pt}
\caption{R-peak detection}
\label{fig1}
\end{figure}

\begin{table*}[]
\centering
\captionsetup{belowskip=-5pt}
\caption{Feature Set}
\label{table:1}
\begin{tabular}{|l|l|}
\hline
\textbf{Features} & \textbf{Description}                                                                     \\ \hline
RMSSD             & Root Mean Square of Successive Normal-to-Normal interval Differences                     \\ \hline
SDSD              & Standard deviation of the time interval between successive NN-intervals                  \\ \hline
SDNN              & Standard deviation of Normal-to-Normal intervals                                         \\ \hline
MEAN HR           & Mean Heart Rate                                                                          \\ \hline
MIN HR            & Minimum Heart Rate                                                                       \\ \hline
MAX HR            & Maximum Heart Rate                                                                       \\ \hline
MEAN RR           & Mean RR Interval                                                                         \\ \hline
NNI\_20           & Number of interval differences of successive NN-intervals greater than 20 ms             \\ \hline
PNNI\_20          & Ratio of NNI\_20 and total number of NN-intervals                                        \\ \hline
NNI\_50           & Number of interval differences of successive NN-intervals greater than 50 ms             \\ \hline
PNNI\_50          & Ration of NNI\_50 and total number of NN-intervals                                       \\ \hline
MEAN\_NNI         & Mean of NN-intervals                                                                     \\ \hline
CVSD              & Coefficient of variation of successive differences equal to ratio of RMSSD and MEAN\_NNI \\ \hline
CVNNI             & Coefficient of variation equal to the ratio of SDNN and MEAN\_NNI                        \\ \hline
\end{tabular}
\end{table*}


All ECG data were downsampled to 250Hz. A low-pass Butterworth filter with a cut-off frequency of 50 Hz was applied on AFDB, CHDB and VFDB to eliminate the ECG baseline wander. The ECG signals were further rescaled to values between 0 and 1 by applying a min-max scaling to reduce inter-dataset discrepancy, caused by the usage of different sensor hardware for ECG signal measurement. Both the AF datasets provide R-peak values. However, during deployment, there will be no clinical expert intervention to mark the R-peak values and the pipeline will receive only ECG data. Hence, R-peaks were automatically extracted for both experimentation and deployment. 

The ECG data from both AFDB and CHDB were segmented with a window length of 5 seconds and an overlap of 1 second. The datasets provide fibrillations as well as flutters and junctional rhythms in their annotations. Each window has 5 seconds of AF annotations and the decision to label an entire window as a positive class depends on a threshold parameter $p$. If the ratio of the number of AF annotations and the total number of annotations (AF and non-AF) exceeds the threshold $p$, then the window segment is classified as an AF signal. The chosen value of $p$ in this study is $0.5$ as adapted from Hongpo Zhang et al. \cite{20}.

For the manual R-peak extraction process, every window segment was passed to a detector that returns the R-peaks contained in that segment. Stationary Wavelet Transform (SWT) \cite{21} detection algorithm was employed in this research. Fig.\ref{fig1} shows the R-peaks after detection by SWT detector.


\section{Pipeline Architecture}
\label{Pipeline Arch}
The pipeline is an end-to-end workflow that takes raw ECG as input and delivers AF prediction. The raw ECG input is passed to a data pre-processing and feature extraction engine which transforms the signal and extracts the necessary features for AF detection. This engine is pipelined to a scalable ML classifier that outputs the prediction. The architecture of this pipeline is shown in Fig. \ref{Pipeline Arch}. The sequential stages of the pipeline are described in-depth in the following subsections. 

\subsection{Feature Extraction Engine}
\label{Feature Extraction Engine}

The Feature Extraction Engine $(\Delta)$ of the pipeline contains 4 sequential stages, namely Baseline Wander Removal $(\alpha)$, Scaling $(\beta)$, R-peak Detection $(\kappa)$ and Feature Extraction $(\lambda)$ which are explained in detail in Section \ref{Dataset preprocessing}. $\gamma$ is the set of raw ECG values with $w$ being window length in seconds, $\nu_s$ the sampling frequency in Hz, $f$ the Heart Rate Variability (HRV) features, and $n$ the variable subset of features discussed in detail, in Section \ref{Feature Selection}. $\Delta$ is depicted in Equation \ref{eq:1}.

\begin{equation}
\label{eq:1}
    \gamma = [\gamma_1 \quad \gamma_2 \quad ... \quad \gamma_{w\times{\nu_s}}] \overset{\Delta}{\longrightarrow} [f_1 \quad f_2 \quad ... \quad f_{n}] 
\end{equation}

\begin{equation}
    \Delta = [\quad \alpha,\quad \beta,\quad \kappa,\quad \lambda \quad ]
\end{equation}

For this study, HRV features were considered, primarily owing to their superior relevance when compared to general non-linear features such as Shannon Entropy, Renyi Entropy and Tsallis entropy \cite{22}. As stated by Gokul H et al. \cite{23}, frequency domain features involve Fast Fourier Transformations which are inherently dilatory to extract. Being statistical in nature, time-domain features can be extracted with less to no delay. $\Delta$ converts R-peaks to 14 time-domain features, as shown in Table \ref{table:1}. 

\subsection{Resource-efficient Classifier}
\label{Resource-efficient Classifier}

The algorithm used, Bonsai \cite{24}, is a tree learner specifically designed for severely resource-constrained devices. Bonsai learns a sparse project matrix that projects the input vector to a lower dimension to reduce data usage. This is implemented in a streaming fashion so as to even accommodate devices with RAMs small enough to not be able to fit a single vector. \par
Bonsai learns a single, shallow, sparse tree that uses enhanced nodes  --- both internal and leaf nodes make non-linear predictions. The overall prediction for a point is the sum of the individual node predictions along the path traversed by the point. Path-based prediction allows Bonsai to accurately learn non-linear decision boundaries while sharing parameters along paths to further reduce model size.\par
Rather than learning the tree node by node in a greedy fashion, all nodes are learned jointly, along with the sparse projection matrix, so as to optimally allocate memory budgets to each node while maximizing prediction accuracy.

\section{Feature Selection and Training}
\label{Feature Selection}
\subsection{Feature Importance}
\label{FeatsImportance}

As the prediction time of the classifier is directly proportional to the number of features used, feature selection was carried out with the aim of finding out the optimal set of features such that the performance of the underlying classifier was maximized while minimizing the number of features selected. This resulted in an optimal feature subset which reduced the computational cost of the classifier and the processing time of $\Delta$.

\begin{figure}[H]
\includegraphics[scale=0.6]{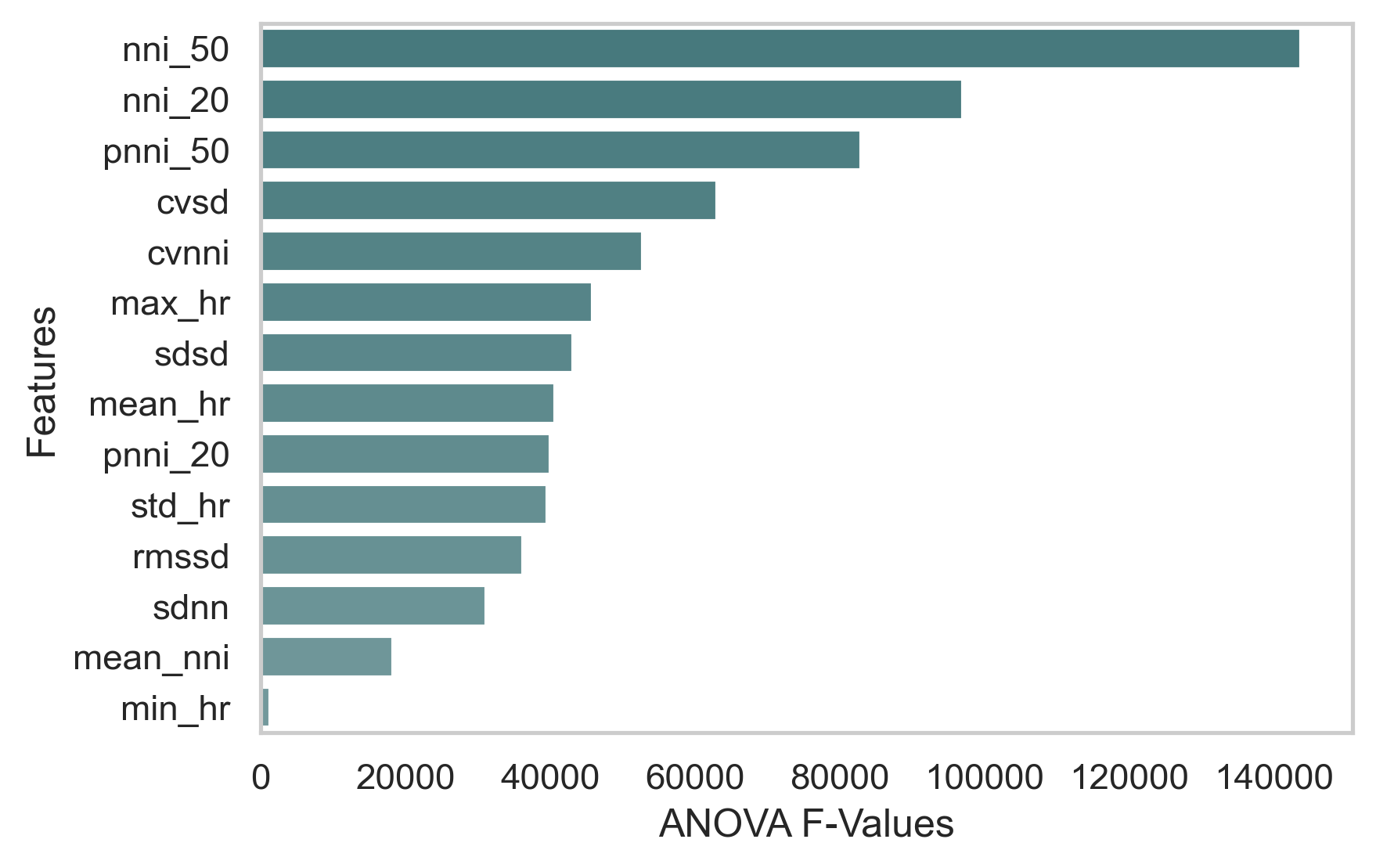}
\captionsetup{belowskip=-10pt}
\caption{CHDB Feature Importance}
\label{2017featureimportance}
\end{figure}

\begin{figure}[H]
\includegraphics[scale=0.6]{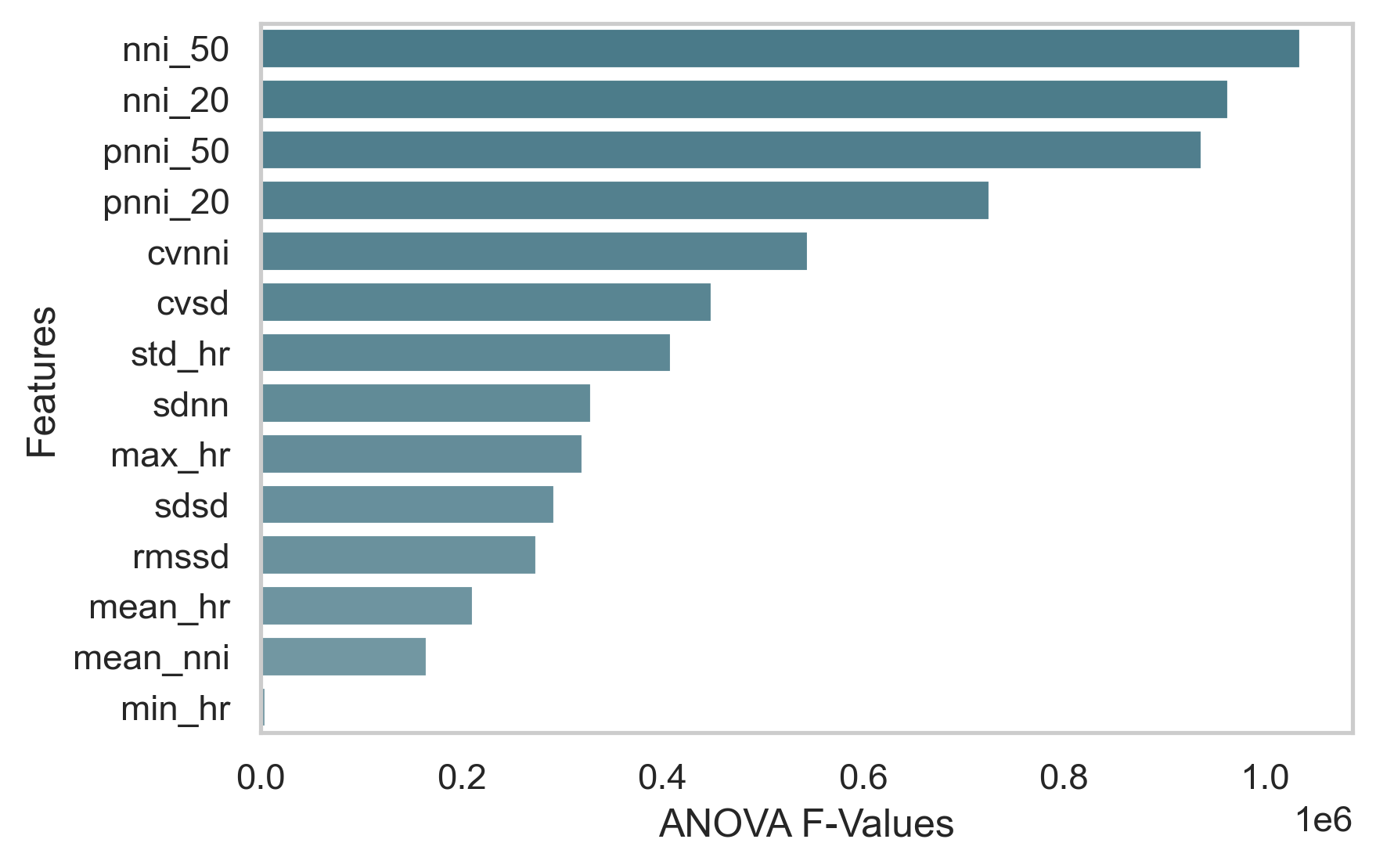}
\captionsetup{belowskip=-10pt}
\caption{AFDB Feature Importance}
\label{afdbfeatureimportance}
\end{figure}

Analysis of Variance (ANOVA) \cite{26} F-test was carried out between each of the independent variables and the predictor variable. An F-statistic or F-value is a measure that determines if the variance between the means of the two populations are significantly different. The higher the value of F-statistic, the higher is the predictive power. The F-values of the extracted features from both the datasets are computed and shown in Fig. \ref{2017featureimportance} and \ref{afdbfeatureimportance}. The features are arranged in descending order of their F-values indicating their feature importance in $F^{Data}$ as shown Equation \ref{eq:2}.

\begin{equation}
\label{eq:2}
\begin{aligned}
F^{Data} = {} & \{f| f_i, F-value(f_i)>F-value(f_{i+1}),  \\
      & 1\leq i \leq 14\}
\end{aligned}
\end{equation}

$Data \longrightarrow A = AFDB, C = CHDB$

\subsection{Training}
\label{Training}

A considerable number of AI systems in the healthcare domain fall short of achieving generalizability, thus clinical applicability. Generalization can be hard due to technical differences between sites (including differences in equipment) as well as variations in local clinical and administrative practices \cite{25}. In order to validate the generalizability, we used 2 datasets that collected data under different clinical settings. We constructed a robust train-test strategy where the models were trained and tested on different datasets. The models were also tested on the same dataset it was trained on to establish a point of comparison. This train-test strategy is elucidated in Tab. \ref{table:2}. 


\begin{equation}
\label{equationA}
    F^A_n = \{ f:f_i | 1\leq i \leq n, f_i \in F^A\} , 1 \leq n \leq 14
\end{equation}
\begin{equation}
\label{equationC}
    F^C_n = \{ f:f_i | 1\leq i \leq n, f_i \in F^C\}, 1 \leq n \leq 14
\end{equation}

Models were trained on several subsets of features to choose the optimal subset based on performance, memory requirements and pipeline inference time. In $T_{AA}$ and $T_{CC}$, the classifier was trained on feature subsets $F_n (n = 4, 6, 8, 10, 12, 14)$, the $n$ most important features chosen using ANOVA as discussed in Section \ref{FeatsImportance}. The best feature subsets for $T_{AA}$ $F^A_{best}$ and $T_{CC}$ $F^C_{best}$ were obtained based on accuracy. $T_{AC}$ and $T_{CA}$ were trained on two subset of features each, as derived from $F^A_{best}\cup F^C_{best}$ and $F^A_{best}\cap F^C_{best}$. A 5-fold cross validation was implemented for $T_{AA}$ and $T_{CC}$. The trained models were deployed onto a Raspberry Pi 3 Model B and the results are discussed in the Section \ref{Results and Analysis}.


\begin{table}[h]
\caption{Train Test combination}
\label{table:2}
\begin{center}
\centering
\begin{tabular}{|c|c|c|c|} 
\cline{3-4}
\multicolumn{1}{l}{}              &               & \multicolumn{2}{c|}{\textbf{Training}}                                                                                                                             \\ 
\cline{3-4}
\multicolumn{1}{l}{}              &               &  \textbf{AFDB(A)}                                                                   & \textbf{CHDB(C)}                                                                    \\ 
\hline
\multirow{2}{*}{\textbf{Testing}} & \textbf{AFDB (A)} & \begin{tabular}[c]{@{}l@{}}$T_{AA}$ \\ \\ Training: AFDB\\ Testing: AFDB\end{tabular} & \begin{tabular}[c]{@{}l@{}}$T_{CA}$ \\ \\ Training: CHDB\\ Testing: AFDB\end{tabular}  \\ 
\cline{2-4}
                                  & \textbf{CHDB(C)} & \begin{tabular}[c]{@{}l@{}}$T_{AC}$\\ \\ Training: AFDB\\ Testing: CHDB\end{tabular}  & \begin{tabular}[c]{@{}l@{}}$T_{CC}$\\ \\ Training: CHDB\\ Testing: CHDB\end{tabular}   \\
\hline
\end{tabular}
\end{center}
\end{table}

\section{Results and Analysis}
\label{Results and Analysis}

All the observations were carried out on a Raspberry Pi 3 Model B and the results were recorded.

\subsection{Metrics} 
The metrics extensively found in literature and throughout this paper are given below by the formulae,

 \begin{equation}
                  Accuracy =  \frac{TP+TN}{TP+TN+FP+FN} 
                    \end{equation}

 \begin{equation}
                    Precision = \frac{TP}{TP+FP} 
                    \end{equation}

\begin{figure*}[h]
\centerline{\includegraphics[scale=0.5]{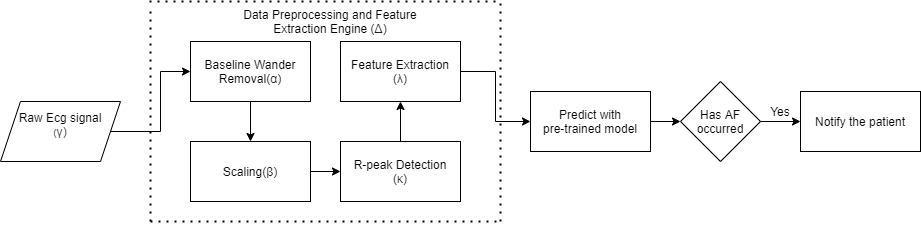}}
\captionsetup{belowskip=-10pt}
\caption{Pipeline Architecture}
\label{deploymentarch}
\end{figure*}
\vspace{5mm}
\begin{table*}[h]
\begin{center}
\caption{Standard ML Models}
\label{table:4}
\begin{tabular}{|c|c|c|c|c|c|c|c|c|c|} 
\hline
\multirow{2}{*}{\textbf{Model}}            & \multirow{2}{*}{\textbf{Data}} & \multirow{2}{*}{\textbf{Accuracy}} & \multicolumn{2}{c|}{\textbf{Precision}} & \multicolumn{2}{c|}{\textbf{Recall}} & \multicolumn{2}{c|}{\textbf{F1 Score}} & \multirow{2}{*}{\begin{tabular}[c]{@{}c@{}}\textbf{Model Size}\\\textbf{ (in kB)}\end{tabular}}  \\ 
\cline{4-9}
                                           &                                &                                    & \textbf{Class 0} & \textbf{Class 1}     & \textbf{Class 0} & \textbf{Class 1}  & \textbf{Class 0} & \textbf{Class 1}    &                                                                                                  \\ 
\hline
\multirow{2}{*}{Decision Tree}             & CHDB                 & 93\%                               & 0.96             & 0.90                 & 0.89             & 0.96              & 0.93             & 0.93                & 378                                                                                              \\ 
\cline{2-10}
                                           & AFDB                           & 88\%                               & 0.89             & 0.88                 & 0.87             & 0.90              & 0.88             & 0.89                & 348                                                                                              \\ 
\hline
\multirow{2}{*}{Random Forest}             & CHDB                 & 83\%                               & 0.82             & 0.85                 & 0.86             & 0.81              & 0.84             & 0.83                & 30699                                                                                            \\ 
\cline{2-10}
                                           & AFDB                           & 87\%                               & 0.88             & 0.87                 & 0.86             & 0.88              & 0.87             & 0.87                & 29388                                                                                            \\ 
\hline
\multirow{2}{*}{Support Vector Classifier} & CHDB                 & 83\%                               & 0.82             & 0.83                 & 0.84             & 0.82              & 0.83             & 0.83                & 563                                                                                              \\ 
\cline{2-10}
                                           & AFDB                           & 89\%                               & 0.88             & 0.89                 & 0.90             & 0.88              & 0.89             & 0.89                & 604                                                                                              \\ 
\hline
\multirow{2}{*}{Extra Trees Classifier}    & CHDB                 & 95\%                               & 0.97             & 0.93                 & 0.93             & 0.97              & 0.95             & 0.95                & 91811                                                                                            \\ 
\cline{2-10}
                                           & AFDB                           & 92\%                               & 0.95             & 0.89                 & 0.88             & 0.95              & 0.91             & 0.92                & 91289                                                                                            \\
\hline
\end{tabular}
\end{center}
\end{table*}

\begin{equation}
                     Recall = \frac{TP}{TP+FN} 
                    \end{equation}

 \begin{equation}
                F1 =  \frac{2*Precision*Recall}{Precision+Recall} = \frac{2*TP}{2*TP+FP+FN}
\end{equation}

$TP$ = $AF$ events classified as $AF$ events

$FP$ = $non$-$AF$ events classified as $AF$ events

$TN$ = $non$-$AF$ events classified as $non$-$AF$ events

$FN$ = $AF$ events classified as $non$-$AF$ events

\vspace{2mm}
The other measurements used to quantify our model's performance on the embedded device include model size and inference time. \textit{Model size} is the memory footprint of the classifier on the embedded device. \textit{Inference time} is termed as the total processing time of $\Delta$.

\subsection{Bonsai Results and Analysis}
\label{Bonsai Results and Analysis}

The most optimal feature subset can be either chosen solely with respect to the model performance or based on the compromise between model performance and size. The reasons for choosing a feature subset for both cases are as follows:

\subsubsection{Best model performance}
\label{bestmodelperformance}

We discerned the best scores for $T_{AA}$ with respect to accuracy, sensitivity and specificity was 91.7\%, 95.1\% and 88.1\% was when $F^A_{12}$ used, respectively. Similarly, for $T_{CC}$, Bonsai delivers the best performance using $F^C_{14}$ with values of 92.3\%, 92.7\% and 91.8\%, respectively. For $T_{AC}$ and $T_{CA}$, we observe that $F^A_{14}\cap F^C_{12}$ gives the best  accuracy (90.3\%, 89.8\%) compared to $T_{AC}\cup T_{CA}$. Since $F^A_{14}\cap F^C_{12}$ uses fewer features than $T_{AC}\cup T_{CA}$ it has lower inference time and model size. Hence, we conclude that $F^A_{14}\cap F^C_{12} = F^C_{12}$ is the most optimal feature subset if performance is given the most importance.

\subsubsection{Best trade-off between model size and performance}

As seen in the below plots, there is a general trend of increase in accuracy with the increase in the number of features. But the increase in accuracy slows down after the top 8 features for both datasets despite a linear increase in model size as seen in Fig. \ref{2017accsize} and \ref{afdbaccsize}. If the optimal feature subset were to be chosen based on the compromise between model performance and size, $F^A_{8}$ or $F^C_{8}$ would be chosen. Considering CHDB, when the number of features used was increased from 4 ($F^c_{4}$) to 8 ($F^C_{8}$) the accuracy increased by 2.0\% while the model size increases by 0.062 KB, but when the number of features used was increased from 8 ($F^C_{8}$) to 14 ($F^C_{14}$) the accuracy increased only by 0.3\% while the model size increases by 0.094 KB, the same trend is seen in AFDB.

\begin{table}[h!]
\centering
\caption{Comparison between CNN+GRU and Bonsai}
\label{Attack Zurich}
\begin{tabular}{|c|c|c|} 
\hline
                             & \textbf{CNN + GRU} & \textbf{Bonsai}  \\ 
\hline
\textbf{ECG window}          & 1 $\times$ 256            & 1 $\times$ 1250      \\ 
\hline
\textbf{Memory [KB]}         & 210.00             & 0.508             \\ 
\hline
\textbf{Accuracy [\%]}       & 86.1               & 92.3              \\ 
\hline
\textbf{Inference [ms]} & 94.8               & 16.9               \\ 
\hline
\end{tabular}
\end{table}

We observe regardless of an increase in the number of features, Bonsai’s inference time remained unchanged. Additionally, when compared to the previous SoA embedded implementation \cite{13}, our classifier’s accuracy was 6.2\% higher, 413$\times$ compact in terms of model size and 5.2$\times$ faster in terms of inference time (Table \ref{Attack Zurich}).


\subsection{Standard ML models}
\label{StandardMLmodels}


To gauge the performance of our classifier, standard ML models such as Decision Trees, Random Forests, Extra Trees and Support Vector Classifier (SVC) were considered as a baseline for comparison. Since Bonsai is a tree-based classifier, three tree-based standard ML models were chosen for a fair comparison. These classifiers were trained on the most optimal feature subset $F^C_{12}$, as derived in Section \ref{bestmodelperformance}. The classification scores and the memory footprint of these models were reported in Table \ref{table:4}. Bonsai was able to outperform the best performing ML model (Extra Trees) around $10^5\times$ in terms of memory footprint with a mere trade-off of 2\% classification accuracy. Bonsai also surpassed the most compact ML model (Decision trees) by a factor of 685$\times$.

\begin{table*}
\centering
\caption{Results of all feature combinations}
\def\arraystretch{1.4}
\label{Results of all feature combinations}
\begin{tabular}{|c|c|c|c|c|c|c|c|c|c|c|c|} 
\hline
\multirow{3}{*}{\textbf{Data}} & \multirow{3}{*}{\begin{tabular}[c]{@{}c@{}}\textbf{Feature}\\\textbf{ Subsets}\end{tabular}} & \multirow{3}{*}{\begin{tabular}[c]{@{}c@{}}\textbf{Accuracy }\\\textbf{ (\%)}\end{tabular}} & \multicolumn{2}{c|}{\multirow{2}{*}{\textbf{Precision (\%)}}}                                                                                         & \multicolumn{2}{c|}{\multirow{2}{*}{\textbf{Recall (\%)}}}                                                                                                              & \multicolumn{2}{c|}{\multirow{2}{*}{\textbf{F1 Score (\%)}~}}                                                                                                                                               & \textbf{Model~\textbf{Size}}~~\textbf{~}                                                                                                             & \begin{tabular}[c]{@{}c@{}}\textbf{Bonsai }\\\textbf{ Inference}\end{tabular}              & \begin{tabular}[c]{@{}c@{}}\textbf{Pipeline }\\\textbf{ Inference}\end{tabular}    \\
                               &                                                                                              &                                                                                             & \multicolumn{2}{c|}{}                                                                                                                                 & \multicolumn{2}{c|}{}                                                                                                                                                   & \multicolumn{2}{c|}{}                                                                                                                                                                                       & \textbf{\textbf{\textbf{\textbf{\textbf{\textbf{\textbf{\textbf{\textbf{\textbf{\textbf{\textbf{\textbf{\textbf{\textbf{\textbf{(KB)}}}}}}}}}}}}}}}} & \textbf{\textbf{Time~}\textbf{\textbf{\textbf{\textbf{\textbf{\textbf{\textbf{(ms)}}}}}}}} & \textbf{Time~\textbf{\textbf{\textbf{\textbf{\textbf{\textbf{\textbf{(ms)}}}}}}}}  \\ 
\cline{4-9}
                               &                                                                                              &                                                                                             & \multicolumn{1}{l|}{\textbf{\textbf{\textbf{\textbf{\textbf{\textbf{\textbf{\textbf{Class 0}}}}}}}}} & \multicolumn{1}{l|}{\textbf{\textbf{Class 1}}} & \multicolumn{1}{l|}{\textbf{\textbf{\textbf{\textbf{\textbf{\textbf{\textbf{\textbf{Class 0}}}}}}}}} & \multicolumn{1}{l|}{\textbf{\textbf{\textbf{\textbf{Class 1}}}}} & \multicolumn{1}{l|}{\textbf{\textbf{\textbf{\textbf{\textbf{\textbf{\textbf{\textbf{Class 0}}}}}}}}} & \multicolumn{1}{l|}{\textbf{\textbf{\textbf{\textbf{\textbf{\textbf{\textbf{\textbf{Class 1}}}}}}}}} &                                                                                                                                                      &                                                                                            &                                                                                    \\ 
\hline
\multirow{6}{*}{$T_{AA}$}      & $F^{A}_{4}$                                                                                  & 89.60                                                                                       & 94.16                                                                                                & 85.95                                          & 84.50                                                                                                & 94.75                                                            & 89.06                                                                                                & 90.13                                                                                                & 0.352                                                                                                                                                & 3.7                                                                                        & 16.5                                                                               \\ 
\cline{2-12}
                               & $F^{A}_{6}$                                                                                  & 90.80                                                                                       & 95.20                                                                                                & 87.18                                          & 85.93                                                                                                & 95.67                                                            & 90.32                                                                                                & 91.22                                                                                                & 0.383                                                                                                                                                & 3.7                                                                                        & 17.0                                                                               \\ 
\cline{2-12}
                               & $F^{A}_{8}$                                                                                  & 91.40                                                                                       & 94.72                                                                                                & 88.51                                          & 87.62                                                                                                & 95.10                                                            & 91.02                                                                                                & 91.68                                                                                                & 0.414                                                                                                                                                & 3.7                                                                                        & 17.2                                                                               \\ 
\cline{2-12}
                               & $F^{A}_{10}$                                                                                 & 91.50                                                                                       & 94.97                                                                                                & 88.47                                          & 87.55                                                                                                & 95.36                                                            & 91.10                                                                                                & 91.78                                                                                                & 0.445                                                                                                                                                & 3.7                                                                                        & 17.6                                                                               \\ 
\cline{2-12}
                               & $F^{A}_{12}$                                                                                 & \textbf{91.70}                                                                              & 94.88                                                                                                & 88.89                                          & 88.07                                                                                                & 95.24                                                            & 91.34                                                                                                & 91.95                                                                                                & 0.477                                                                                                                                                & 3.7                                                                                        & 17.7                                                                               \\ 
\cline{2-12}
                               & $F^{A}_{14}$                                                                                 & 91.60                                                                                       & 94.73                                                                                                & 88.85                                          & 88.05                                                                                                & 95.12                                                            & 91.27                                                                                                & 91.87                                                                                                & 0.508                                                                                                                                                & 3.7                                                                                        & 18.1                                                                               \\ 
\hline
\multirow{6}{*}{$T_{CC}$}      & $F^{C}_{4}$                                                                                  & 89.95                                                                                       & 91.34                                                                                                & 88.69                                          & 88.31                                                                                                & 91.62                                                            & 89.79                                                                                                & 90.12                                                                                                & 0.352                                                                                                                                                & 3.7                                                                                        & 16.9                                                                               \\ 
\cline{2-12}
                               & $F^{C}_{6}$                                                                                  & 91.60                                                                                       & 92.18                                                                                                & 91.03                                          & 90.90                                                                                                & 92.28                                                            & 91.53                                                                                                & 91.64                                                                                                & 0.383                                                                                                                                                & 3.7                                                                                        & 16.9                                                                               \\ 
\cline{2-12}
                               & $F^{C}_{8}$                                                                                  & 91.97                                                                                       & 92.54                                                                                                & 91.42                                          & 91.30                                                                                                & 92.63                                                            & 91.91                                                                                                & 92.02                                                                                                & 0.414                                                                                                                                                & 3.7                                                                                        & 17.4                                                                               \\ 
\cline{2-12}
                               & $F^{C}_{10}$                                                                                 & 92.05                                                                                       & 92.03                                                                                                & 92.06                                          & 92.08                                                                                                & 92.03                                                            & 92.05                                                                                                & 92.04                                                                                                & 0.445                                                                                                                                                & 3.7                                                                                        & 17.5                                                                               \\ 
\cline{2-12}
                               & $F^{C}_{12}$                                                                                 & 92.09                                                                                       & 92.31                                                                                                & 91.88                                          & 91.83                                                                                                & 92.34                                                            & 92.07                                                                                                & 92.11                                                                                                & 0.477                                                                                                                                                & 3.7                                                                                        & 17.6                                                                               \\ 
\cline{2-12}
                               & $F^{C}_{14}$                                                                                 & \textbf{92.25}                                                                              & 92.64                                                                                                & 91.87                                          & 91.79                                                                                                & 92.71                                                            & 92.21                                                                                                & 92.29                                                                                                & 0.508                                                                                                                                                & 3.7                                                                                        & 18.0                                                                               \\ 
\hline
\multirow{2}{*}{$T_{AC}$}      & $F^A_{12}\cap F^C_{14}$                                                                      & 90.30                                                                                       & 88.90                                                                                                & 90.50                                          & 90.70                                                                                                & 88.70                                                            & 89.80                                                                                                & 89.60                                                                                                & 0.477                                                                                                                                                & 3.7                                                                                        & 16.3                                                                               \\ 
\cline{2-12}
                               & $F^A_{12}\cup F^C_{14}$                                                                      & 89.70                                                                                       & 89.90                                                                                                & 88.60                                          & 88.40                                                                                                & 90.10                                                            & 89.10                                                                                                & 89.30                                                                                                & 0.508                                                                                                                                                & 3.7                                                                                        & 16.4                                                                               \\ 
\hline
\multirow{2}{*}{$T_{CA}$}      & $F^A_{12}\cap F^C_{14}$                                                                      & 89.80                                                                                       & 95.20                                                                                                & 85.50                                          & 83.80                                                                                                & 95.80                                                            & 89.10                                                                                                & 90.30                                                                                                & 0.477                                                                                                                                                & 3.7                                                                                        & 16.6                                                                               \\ 
\cline{2-12}
                               & $F^A_{12}\cup F^C_{14}$                                                                      & 89.60                                                                                       & 95.80                                                                                                & 84.90                                          & 82.80                                                                                                & 96.30                                                            & 88.80                                                                                                & 90.20                                                                                                & 0.508                                                                                                                                                & 3.7                                                                                        & 16.9                                                                               \\
\hline
\end{tabular}
\end{table*}

\raggedbottom
\subsection{Pipeline Analysis}

\begin{figure}[H]
\includegraphics[scale=0.55]{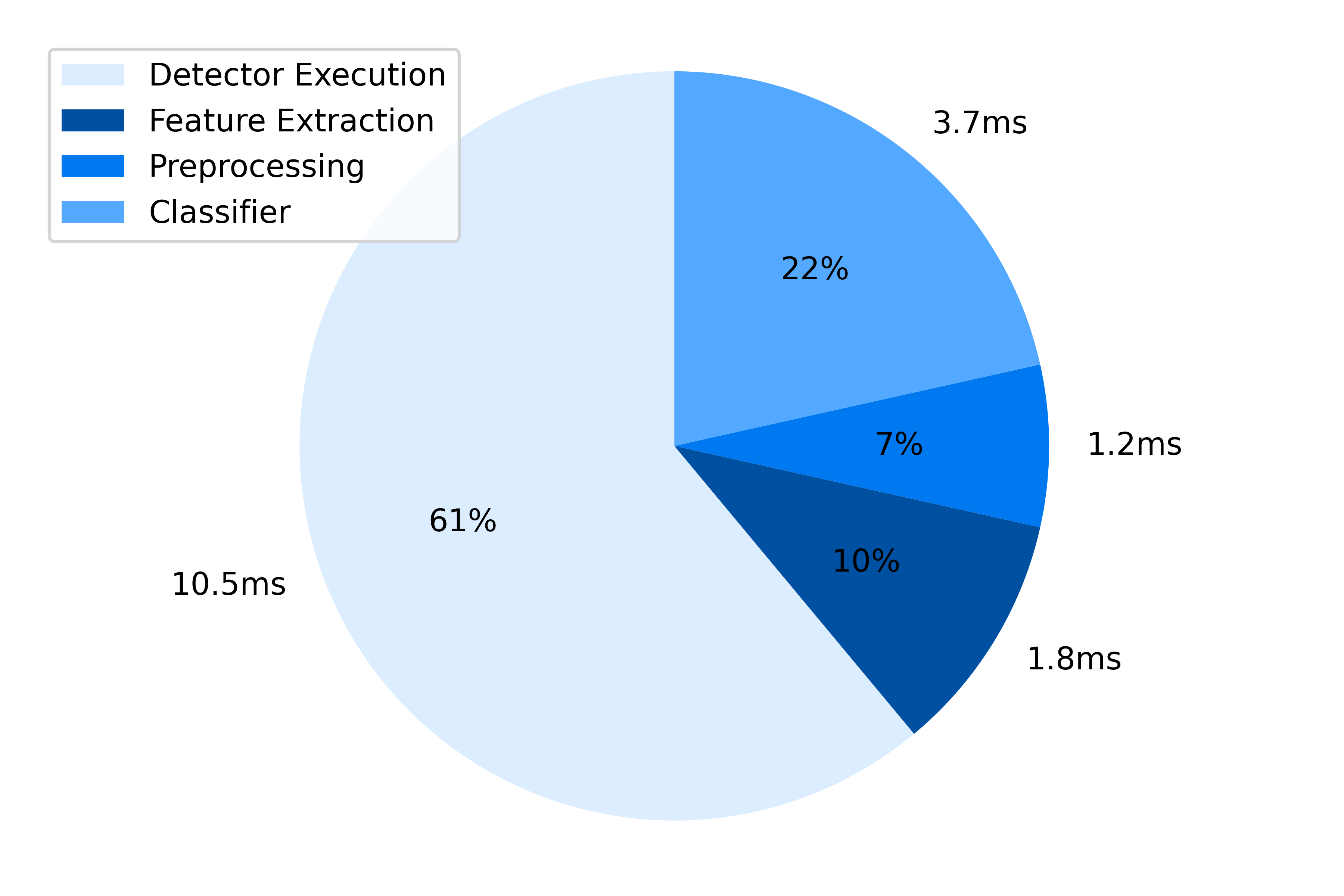}
\captionsetup{belowskip=-10pt}
\caption{Distribution of computation time}
\label{piechart}
\end{figure}
\begin{figure}[H]
\includegraphics[scale=0.55]{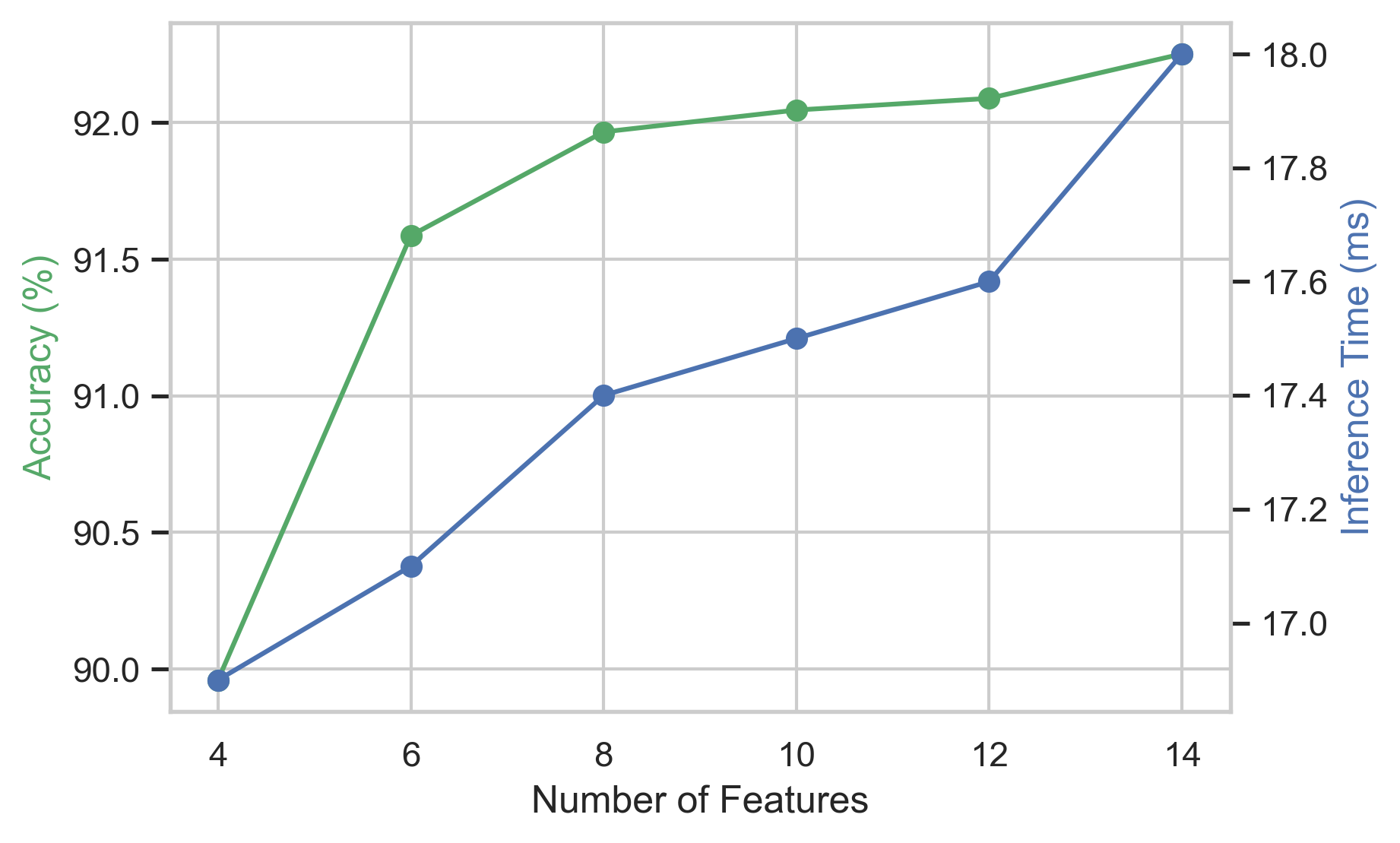}
\captionsetup{belowskip=-10pt}
\caption{CHDB Accuracy vs Inference}
\label{2017accinf}
\end{figure}
\begin{figure}[H]
\includegraphics[scale=0.55]{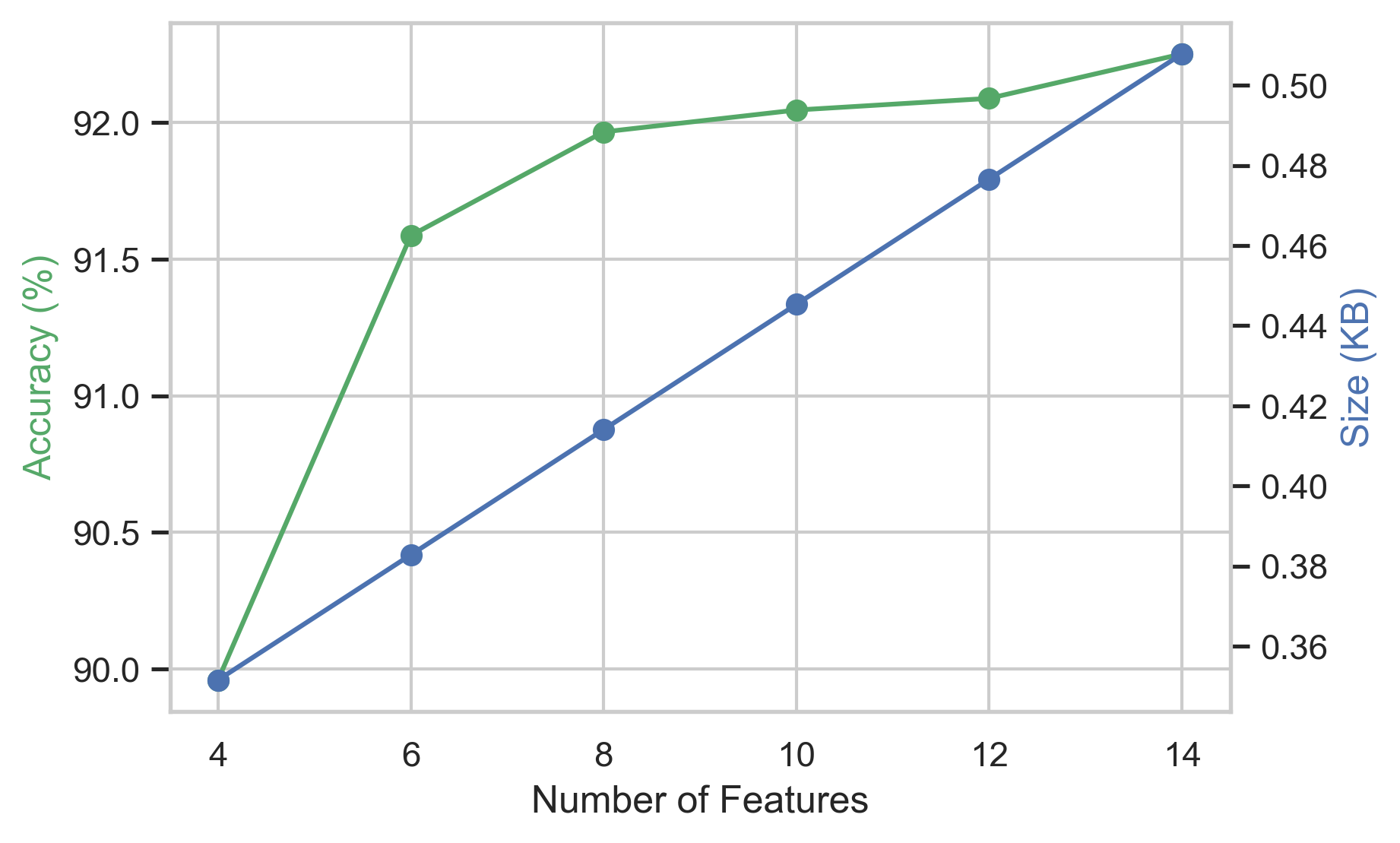}
\captionsetup{belowskip=-10pt}
\caption{CHDB Accuracy vs Size}
\label{2017accsize}
\end{figure}
\begin{figure}[H]
\includegraphics[scale=0.55]{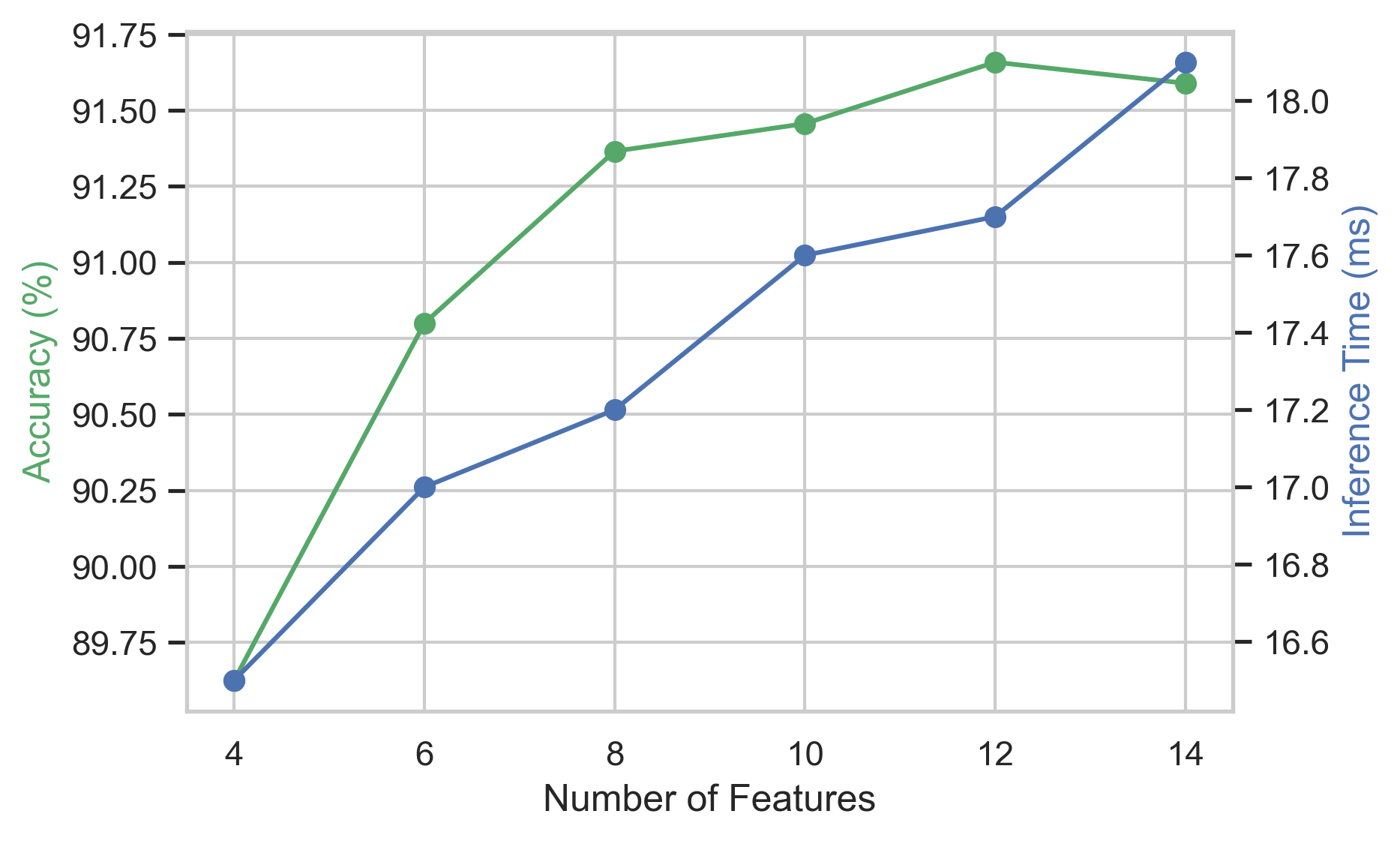}
\captionsetup{belowskip=-10pt}
\caption{AFDB Accuracy vs Inference}
\label{afdbaccinf}
\end{figure}
\begin{figure}[H]
\includegraphics[scale=0.55]{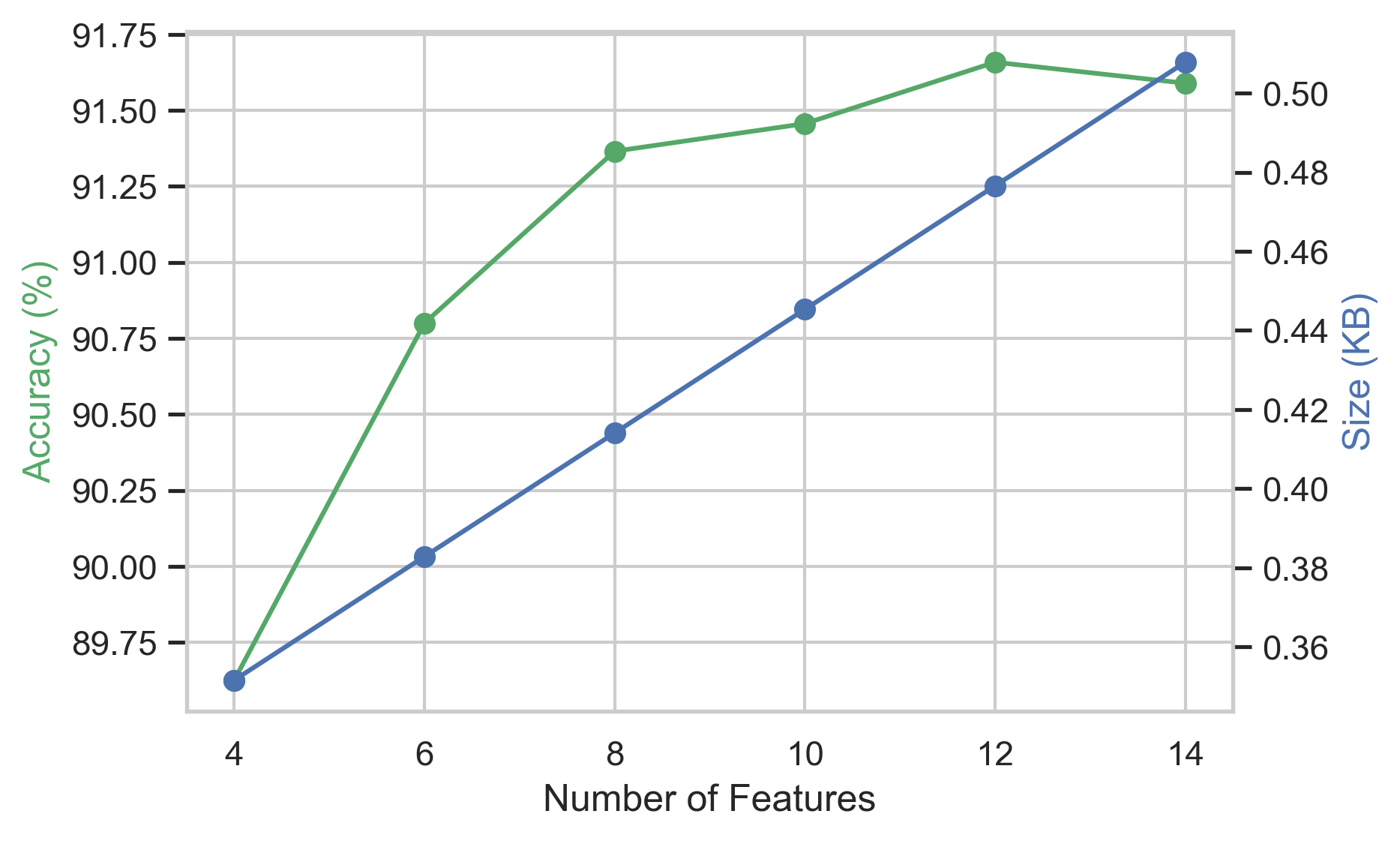}
\captionsetup{belowskip=-10pt}
\caption{AFDB Accuracy vs Size}
\label{afdbaccsize}
\end{figure}

\vspace{20mm}
As inferred from the Fig. \ref{piechart}, pipeline inference time is largely occupied (61\%) by R-peak detector execution time. It is observed from Fig. \ref{2017accinf} and \ref{afdbaccinf} that pipeline inference time increases with growth in the number of features, however, that increase is insignificant (3\%-10\%).

In order to prove the versatility of our pipeline for the detection of other arrhythmias, the pipeline was trained and tested on Ventricular Fibrillation Database as well. The results of which are accuracy - 89\%, F1 score - 86\%, precision - 80\% and recall - 93\%. These scores validate that the proposed pipeline performs well for arrhythmia detection in general.

\vspace{5mm}

\section{Conclusion}
In this research, we present a novel clinically viable end-to-end pipeline for AF detection. The proposed work outperforms the present SoA embedded implementation with respect to the accuracy, inference time and memory footprint. On average, there was a 6\% absolute increase in overall accuracy, while the inference times were observed to be 5.2$\times$ lesser and the model size was 403$\times$ times smaller when compared to the state-of-the-art model. 

The pipeline thus facilitates accurate AF detection, with a significant reduction in latency and memory footprint. Hence they can be pushed onto ultra-edge devices to make real-time on-device predictions. In the future, our end-to-end pipeline could eventually be integrated into a wearable device that acquires and processes ECG signals for arrhythmia detection in real-time.

\printbibliography
\nocite{*}

\end{document}